\documentclass[article,pdftex,10pt,a4paper,journal]{IEEEtran}

\pdfoutput=1
\usepackage{mathtools}
\usepackage{cite}
\usepackage{multirow}
\usepackage{float}
\usepackage[pdftex]{graphicx}
\usepackage{subfigure}
\usepackage{multicol}
\usepackage{paralist}
\usepackage{bm}
\usepackage{amsfonts}
\usepackage{url}
\usepackage{multirow}
\usepackage{color}
\usepackage{amsmath}
\usepackage{theorem}

\newtheorem{lemma}{Lemma}

\usepackage{algorithm}
\usepackage{algorithmicx}
\usepackage{algpseudocode}
\usepackage{ifpdf}

\begin{document}

\pdfoutput=1
\ifpdf
\twocolumn[
\begin{@twocolumnfalse}
\begin{center}
``This paper has been submitted to the IEEE transactions on cybernetics."\par
\end{center}
``$@$2019 IEEE. Personal use of this material is permitted. Permission from IEEE must be obtained for all other uses, in any current or future media, including reprinting/republishing this material for
advertising or promotional purposes, creating new collective works, for resale or redistribution to
servers or lists, or reuse of any copyrighted component of this work in other works."
\end{@twocolumnfalse}
]

\newpage
\title{Incremental Reinforcement Learning --- a New Continuous Reinforcement Learning Frame Based on Stochastic Differential Equation methods}

\author{Tianhao Chen,~Limei Cheng,~Yang Liu,~Wenchuan Jia and Shugen Ma,~\IEEEmembership{Fellow, IEEE}
\thanks{Tianhao Chen, Limei Cheng, Wenchuan Jia and Shugen Ma are with School of Mechatronic Engineering and Automation, Shanghai University, Shanghai, P. R. China. Shugen Ma is also with Department of Robotics, Ritsumeikan University, Japan. }
\thanks{Yang Liu is with American University.}
\thanks{Wenchuan Jia is the corresponding author (e-mail: lovvchris@shu.edu.cn).}}

\markboth{IEEE TRANSACTIONS ON CYBERNETICS}%
{Shell \MakeLowercase{\textit{et al.}}: Bare Demo of IEEEtran.cls for IEEE Journals}

\maketitle
\begin{abstract}
Continuous reinforcement learning such as DDPG and A3C are widely used in robot control and autonomous driving. However, both methods have theoretical weaknesses. While DDPG cannot control noises in the control process,  A3C does not satisfy the continuity conditions under the Gaussian policy. To address these concerns, we propose a new continues reinforcement learning method based on stochastic differential equations and we call it Incremental Reinforcement Learning (IRL).  This method not only guarantees the continuity of actions within any time interval, but controls the variance of actions in the training process. In addition, our method does not assume Markov control in agents' action control and allows agents to predict scene changes for action selection. With our method, agents no longer passively adapt to the environment. Instead, they positively interact with the environment for maximum rewards.
\end{abstract}


\IEEEpeerreviewmaketitle

\section{Introduction}

\IEEEPARstart{D}{eep} reinforcement learning (DRL) is the love child of deep learning (DL) and reinforcement learning (RL). Through extensive data training and pattern recognition, DL methods largely improve the efficiency of feature identification. In order to achieve the maximum reward for characterizing the environment, the optimal state action function from Bellman optimal equation are adopted in newer RL methods. DRL method combines the perceptive ability of DL with the decision-making ability of RL and thus offers strong versatility and direct control from input to output \cite{SJ:15}. These end-to-end learning systems have achieved great success in dealing with complex tasks.\par

As shown in Figure 1, major RL methods are mostly Value based, Policy based and Actor-critic based. Initially,  high-dimensional input data could not be processed directly due to lack of training data and computational power. We rely on Deep Neural Networks (DNN) to reduce the dimensionality of the input data for real world tasks. The well-known Q-learning algorithm in RL, proposed by Watkins \cite{CJ:89}, is a very effective learning algorithm that are widely used in traditional RL\cite{RH:96,WD:02, MA:08}. Various RL algorithms are developed from Q-learning such as Deep Q-network (DQN) algorithm\cite{MV:13, MV:15} (which combines the convolutional network with the traditional Q-learning algorithm in RL) and Deep Dual Q-network (DDQN) algorithm\cite{VH:16}. \par

Although DQN and DDQN have achieved success in solving high-dimensional RL problems, their actions are based on discrete time and thus cannot be applied to continuous cases. To extend Q-learning to continuous state and motion spaces, Gu et al. \cite{SG:16} propose a relatively simple solution called Normalized Advantage Functions (NAF). Although NAF can theoretically perform continuous action output, in fact, it is difficult to apply because of its large computational complexity and training complexity. The Deep Deterministic Policy Gradient (DDPG) based on the Actor-Critic (AC) framework \cite{LT:15} uses policy gradient to perform continuous operation so it can be applied to broader tasks such as robotic control. Nevertheless, DDPG can easily expand rewards and randomness of the policy in complex and noisy situations. The Asynchronous Advantage Actor-Critic (A3C) algorithm \cite{MV:16} is widely used in tasks with both discrete and continuous action space, and arguably achieves the best results concerns. However, the delay in policy update leads to instability.  The Trust Region Policy Optimization (TRPO) \cite{SJ:15} improves the performance of DDPG by introducing a surrogate objective function and a KL divergence constraint, which guarantees a long-term reward without reductions. The Proximal Policy Optimization (PPO) \cite{JS:17} algorithm is built on TRPO algorithm but as a policy gradient algorithm, it has the limitation on the policy gradient algorithm -- the parameters are updated along the direction of the policy gradient despite that PPO optimize TRPO by modifying the objective function.\par

\begin{figure}[ht]
\centering
\includegraphics[width=9cm]{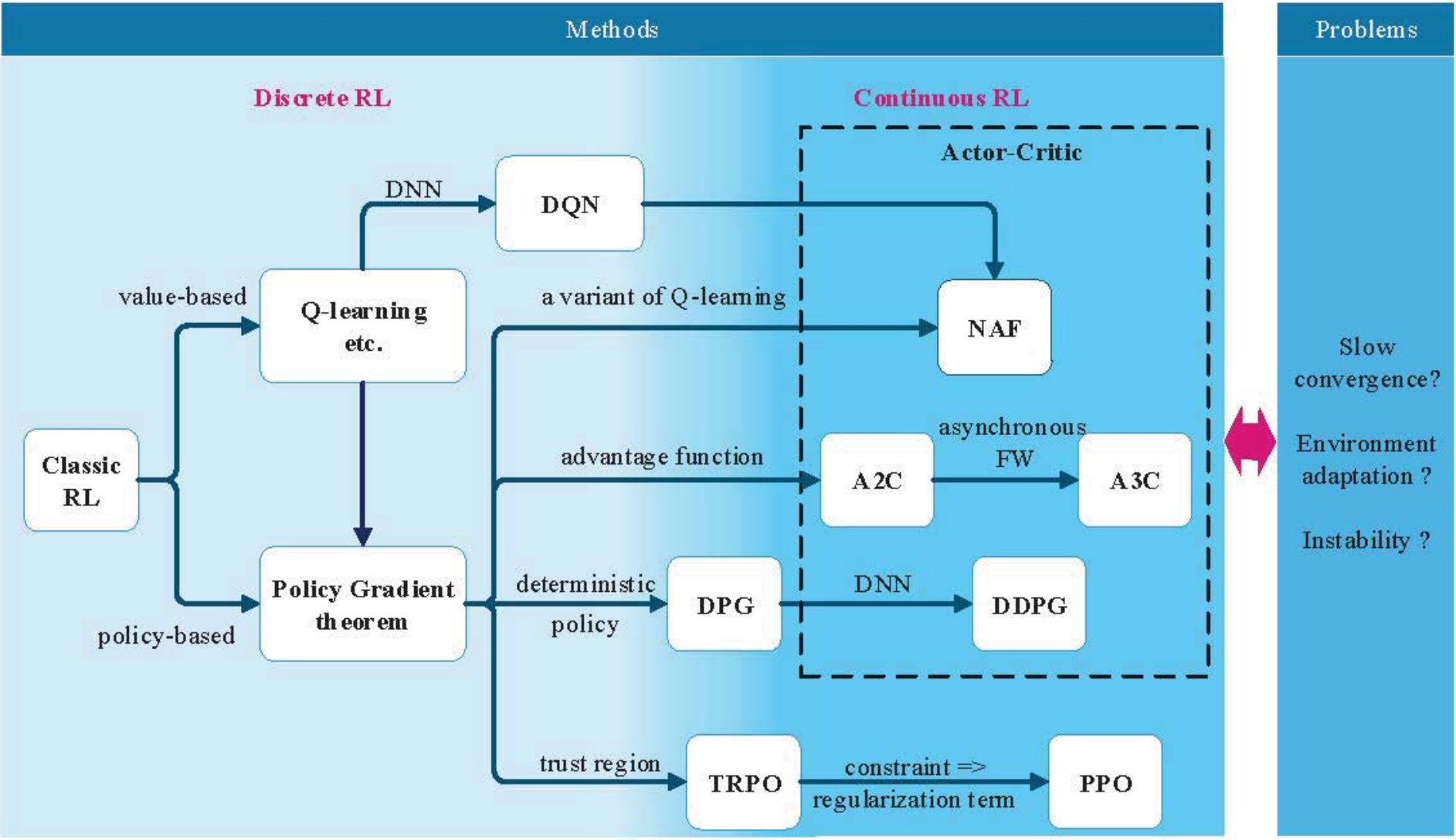}
\caption{Development of reinforcement learning methods}
\label{fig: methods}
\end{figure}

DRL has been widely used in simulation \cite{FQ:14}, robotic control \cite{KJ:13}, optimization and scheduling \cite{WY:05, IE:08}, video games \cite{TG:94, KL:06} and  robotics \cite{SD:16, DY:16, GS:16, HS:16, AM:16, OJ:15, JC:15}. For example, Zhang et al. \cite{ZF:15} show that the trained depth Q-network system can use external visual observations to perform target contact learning so camera images could be replaced by the network composite images. \par

Nevertheless, DRL still has many problems in industrial applications. The most important one of RL comes from its instability, which is also one of the most fundamental differences from the traditional adaptive control. The problem is related to the agent's wide-area exploration capability. Reducing the instability in the RL algorithm while ensuring and encouraging agents to explore during the training process is one of the biggest challenges in reinforcement learning. It is also considered an inherent weakness of classical continuous RL algorithms.\par

For example in DDPG, the action policy $a_t$ is generally $a_t = f^{\theta}(s_t)+\mathfrak{N}_t$, where $s_t$ is environmental state, $\theta$ is the parameter, $\mathfrak{N}_t$ is the white noise or the Uhlenbeck-Ornstein random process. When the Uhlenbeck-Ornstein random process is chosen (which expression is $dX_t=\theta\left(\mu-X_t\right)dt+\sigma dB_t$, $\theta$,$\mu$,$\sigma>0$), despite the continuity condition might be satisfied, we cannot guarantee the variance of the action output. When examining the A3C algorithm under the Gaussian strategy, its action strategy is generally $a_t=\mu^\theta\left(s_t\right)+\sigma^\theta\left(s_t\right)\cdot\mathfrak{N}_t$, where $\mathfrak{N}_t$ is white noise. Although the accuracy can be controlled according to Kolmogorov continuity theorem, A3C's action policy is not continuous, and it cannot find a continuous correction (Appendix A.1). A3C has to use the policy gradient principle to update the network parameters. If $\mathfrak{N}_t$ and DDPG adopt the Uhlenbeck-Ornstein stochastic process, the distribution $\pi\left(a_t|s_t\right)$ will be very difficult to draw, and it is not suitable to use $\nabla_\theta J\left(\theta\right)=E\left[\nabla_\theta\pi\left(a_t | s_t\right)\cdot Q\left(s_t,a_t\right)\right]$ to update the parameters, where $Q\left(s_t,a_t\right)$ is the Q function in RL. In addition, because of the need for discretizing these formulas in practice, the above methods cannot control the size of the time interval. Therefore, when the environmental state is abrupt due to the Markov property, the range of motion changes can reach an unintended situation, which may cause damage to many RL applications such as robotics.\par
Traditional RL methods have difficulties in balancing the uncertainties caused by continuity and variance control. This is also a drawback of the Markov control in RL. Solutions might be found in A3C's action strategy. In stochastic differential equations (SDE) \cite{BK:06}, the white noise $\mathfrak{N}_t$ is often seen as the ``derivative" of the Brownian motion $B_t$ \cite{HW:18}, so the total derivative of A3C algorithm can be written as: $\eta_tdt=\mu^\theta\left(s_t\right)dt+\sigma^\theta\left(s_t\right)\cdot dB_t$. This gives a starting point for our research. Without taking direct action in the Markov controlled output, we focus on incremental changes denoted as $da_t=\eta_tdt$. Then we have the SDE: $da_t=\mu^\theta\left(s_t\right)dt+\sigma^\theta\left(s_t\right)\cdot dB_t$, $a_t\in\mathbb{R}^m$.  In fact, this expression is quite common in practice. For example, a robot joint is usually controlled by its angle rather than the angular velocity. The problem is difficult to solve in traditional RL because the environmental state can hardly be predicted and consequently agents can only adapts to the environment. If the form of the change $s_t$ is not clear, it is almost impossible to make any effective description of the above SDE. Hence, it is necessary to make corresponding estimates of the environmental state.\par
In traditional RL theory, the distribution of the environmental state at the next moment is based on the current environmental state and the agent's action strategy (causing environmental changes), but in reality the environment state evolves continuously. Is it possible to construct a SDE similar to action strategy for the environmental state $s_t$? The answer is
$$ ds_t=f^{\theta_p}\left(s_t,a_t\right)dt+g^{\theta_p}\left(s_t,a_t\right)\cdot dB_t,\ s_t\in\mathbb{R}^n $$
where $f:\mathbb{R}^{n+m}\rightarrow\mathbb{R}^n$,$g:\mathbb{R}^{n+m}\rightarrow\mathbb{R}^{n\times n}$ are functions for environmental change (the network is used instead in the actual process); $\theta_p$ is the parameter set in the equation; $B_t$ is the n-dimensional Brownian motion (each component is independent). This SDE shows that the environment is not only continuously changing, but also depends on the current state of the environment and actions, which satisfies both requirements for a ideal environment state dynamic.\par
It is also necessary to modify the strategy followed by the action policy,
$$da_t=\mu^{\theta_v}\left(s_t,a_t\right)dt+\sigma^{\theta_v}\left(s_t,a_t\right)\cdot d{\widetilde{B}}_t,\ a_t\in\mathbb{R}^m$$
where $\mu:\mathbb{R}^{n+m}\rightarrow\mathbb{R}^m$, $\sigma:\mathbb{R}^{n+m}\rightarrow\mathbb{R}^{m\times m}$ is the equation followed by environmental change (the network is used instead in the actual process); $\theta_p$ is the parameter set in the equation; ${\widetilde{B}}_t$ is the n-dimensional Brownian motion, it is also independent of $B_t$ (each component is independent). Such modifications are reasonable. For example, when a robot makes an action under reinforcement learning, its current action needs to be considered for determining the amount of change. When the allowable angle of the joint rotation is large, the natural motion amplitude might be set larger; when it is close to the critical point, the allowable motion change range is smaller.\par
The two core SDEs in this paper are therefore:
$$ds_t=f^{\theta_p}\left(s_t,a_t\right)dt+g^{\theta_p}\left(s_t,a_t\right)\cdot dB_t,  s_t\in\mathbb{R}^n$$
$$da_t=\mu^{\theta_v}\left(s_t,a_t\right)dt+\sigma^{\theta_v}\left(s_t,a_t\right)\cdot d{\widetilde{B}}_t,  a_t\in\mathbb{R}^m$$
Although the two equations are observed separately, it must be a non-time-aligned SDE. If $Y_t=\left(s_t,a_t\right)$, the two equations can be merged into:
$$
dY_t=
\left(\!\begin{array}{c}
	f^{\theta_p}\left(Y_t\right)\\
	\mu^{\theta_v}\left(Y_t\right)
\end{array}\!\right)
d_t+ \left(\!\!\begin{array}{cc}
    g^{\theta_p}\left(Y_t\right)\!\!&\!\!0\\
    0\!\!&\!\!\sigma^{\theta_v}\left(Y_t\right)
\end{array}\!\!\right)\cdot
\left(\!\begin{array}{c}
dB_t\\
d{\widetilde{B}}_t
\end{array}\!\right) $$
or:
$$dY_t=F^{\left(\theta_p,\theta_v\right)}\left(Y_t\right)dt+G^{\left(\theta_p,\theta_v\right)}\left(Y_t\right)\cdot d{\overline{B}}_t\ $$
where $Y_t\in\mathbb{R}^{n+m}$.
Then $Y_t$ follows I\^{t}o diffusion. This has brought great convenience to our discussion and it is also the core concept of this paper. In fact, this is aligned with the reality - the action of the agent is always interactive with the changing environment. Besides, the Q function in RL is often used to evaluate the future benefit of the current action and state, that is, $Q^\theta(s,a)$, where $\theta$ is the parameter set of the Q-value network. This expression can be changed to $Q^\theta\left(y\right)$ under the current framework.\par

In summary, the entire Incremental Reinforcement learning framework consists of three main parts (and its network) — (1) Environmental state estimator (ESE); (2) Action policy generator (APG); (3) Value estimator (VE) (or Q function). The rest of the paper is arranged as following: Section \ref{sec:The Demonstration of Theory} introduces our incremental reinforcement learning framework; Section \ref{sec:Technical Frame} shows how to update the parameters in our proposed frame. Section \ref{sec:Experiment Results} describes the experimental results; Section V concludes. Proofs for some steps are left in appendix\ref{app:theorem}.\par

\section{Theory}\label{sec:The Demonstration of Theory}
This section consists of four parts: the regulation terms derived from the existence and uniqueness of stochastic differential equations and the description of - ESE, APG and VE. Most of the knowledge about stochastic differential equations is elaborated in \cite{WC:16} so we do not repeat their proof. \par
As mentioned in the introduction, the entire system of IRL is in the form:
$$ds_t =f^{\theta_p}(s_t,a_t)dt+g^{\theta_p}(s_t,a_t)\cdot dB_t, s_t\in \mathbb{R}^n \eqno(1)$$
$$da_t =\mu^{\theta_v}(s_t,a_t)dt+\sigma^{\theta_v}(s_t,a_t)\cdot d\tilde{B}_t, a_t\in \mathbb{R}^m \eqno(2)$$
where $f:\mathbb{R}^{m+n}\to\mathbb{R}^{n};\, \mu:\mathbb{R}^{m+n}\to\mathbb{R}^{m};\, g:\mathbb{R}^{m+n}\to\mathbb{R}^{n\times n};\, \sigma:\mathbb{R}^{m+n}\to\mathbb{R}^{m\times m}$ are all measuable functions. Suppose $Y_t=(s_t,a_t)$, then the stochastic differential equations of the whole system can be transformed into I\^{t}o diffusion equation:
$$dY_t =F^{(\theta_p,\theta_v)}(Y_t)dt+G^{(\theta_p,\theta_v)}(Y_t)\cdot d\overline{B}_t,Y_t\in \mathbb{R}^{m+n} \eqno(3)$$
where $ 
\overline{B}_t=
\left(\begin{array}{c}
	B_t\\
	\tilde{B}_t
\end{array}\right)
$ is a (n+m)-dimensional Brownian motion, and:\\
$$ 
F^{(\theta_p,\theta_v)}(Y_t)=
F^{(\theta_p,\theta_v)}(s_t,a_t)=
\left(\begin{array}{c}
f^{\theta_p}(s_t,a_t)\\
\mu^{\theta_v}(s_t,a_t)
\end{array}\right)
\eqno(4)$$
$$ G^{(\theta_p,\theta_v\!)}(Y_t)\!=\!
G^{(\theta_p,\theta_v\!)}(s_t,\!a_t\!)\!=\!
\left(\!\!\!\!\begin{array}{cc}
	g^{\theta_p}(s_t,a_t)\!\!&\!\!0\\
	0\!\!&\!\!\sigma^{\theta_v}(s_t,a_t)
\end{array}\!\!\!\!\right)\eqno(5)$$
\par
As the basic model of IRL, the discussion of the I\^{t}o diffusion equation of the above system will run through the entire algorithm framework.

\subsection{The regulation terms derived from the existence and uniqueness of stochastic differential equations}
Before discussing the other three core networks, it is necessary to discuss the existence and uniqueness of equation (3). It is obvious that if the solution of the diffusion equation does not exist, it would be hard to get a good interaction between the environment state and agent's action through network training. In order to ensure that the solution not only uniquely exist, but also satisfy the continuity condition, we use the existence and uniqueness theorem of the solution to the stochastic differential equation described below:
\begin{lemma}{Existence and uniqueness of solutions of SDE} 
\label{rem: Existence and uniqueness of solutions of SDE}
\mbox{}\par
\rm Suppose $T>0,b(\cdot,\cdot):[0,T]\times \mathbb{R}^n\to\mathbb{R}^n,\sigma(\cdot,\cdot):[0,T]\times \mathbb{R}^n\to\mathbb{R}^{n\times m}$ are all measuable function. If there exists constant C and D that:
\begin{flalign*}
&(i)|b(t,x)|+|\sigma(t,x)|\leq C(1+|x|),x\in\mathbb{R}^n,t\in[0,T]&\\
&(ii)|b(t,x)-b(t,y)|+|\sigma(t,x)-\sigma(t,y)|\leq D|x-y|,&\\
&\ \ \ \ x,y\in\mathbb{R}^n,t\in[0,T]&
\end{flalign*}
where $|\sigma|^2=\sum{|\sigma_{ij}|^2}$ and $Z$ is a random varible that is independent of the $\sigma$ algebra $\mathcal{F}_{\infty}^{(m)}$ and generated by $\left\{B_s;s\leq t\right\}$. It also satisfies $E[|Z|^2]<\infty$, then the stochastic differential equation:
$$dX_t=b(t,X_t)dt+\sigma (t,X_t)\cdot dB_t$$
has a unique continuous solution $X_t(\omega)$, which is adapted to the filtration $\mathcal{F}_t^Z:=\sigma$\{$Z$ and $B_s;s\leq t$\} and satisfies $E[\int_{0}^{T}{|X_t|^2dt}]<\infty$.
\end{lemma} \par
Acording to lemma \ref{rem: Existence and uniqueness of solutions of SDE}, it is obvious that if the solution of the stochastic differential equation (3) exists, the functions $F^{(\theta_p,\theta_v)}(y),G^{(\theta_p,\theta_v)}(y)$ have to satisfy
\begin{flalign*}
&(i)|F^{(\theta_p,\theta_v)}(y)|+|G^{(\theta_p,\theta_v)}(y)|\leq C(1+|y|),y\in\mathbb{R}^{m+n},&\\
&\ \ \ \ t\in[0,T]&\\
&(ii)|F^{(\theta_p,\theta_v)}(x)-F^{(\theta_p,\theta_v)}(y)|+|G^{(\theta_p,\theta_v)}(x)-G^{(\theta_p,\theta_v)}(y)|&\\
&\ \ \ \ \leq D|x-y|,x,y\in\mathbb{R}^{m+n},t\in[0,T]&
\end{flalign*}
The condition $(i)$ is automatically satisfied because process $Y_t$ is a diffusion process. For condition $(ii)$, it means that $F^{(\theta_p,\theta_v)}$ and $G^{(\theta_p,\theta_v)}$ have to satisfy Lipschitz continuous condition, ($f^{\theta_p},g^{\theta_p},\mu^{\theta_v}$ and $ \sigma^{\theta_v}$). It also means that it is necessary to limit the gradient of the parameters $(\theta_p,\theta_v)$, so a gradient penalty should be added in the objective functions as a regularizer. Hence, two regularizers are constructed 
$$J_{Lip}(\theta_p)=\lambda_1 (\Vert\partial f^{\theta_p}(s,a)\Vert_2+\Vert\partial g^{\theta_p}(s,a)\Vert_2-D_1)^+\eqno(6)$$
$$J_{Lip}(\theta_v)=\lambda_2 (\Vert\partial \mu^{\theta_v}(s,a)\Vert_2+\Vert\partial \sigma^{\theta_v}(s,a)\Vert_2-D_2)^+\eqno(7)$$
where the constants $D_1,D_2$ are the upper bounds of the gradients of the input pair of the network in ESE and APG; $\lambda_1,\lambda_2$ are coefficients for regularizers; $\partial$ is the differential operator (or divergence  operator); $\Vert\cdot\Vert_2$ is 2-norm; and $(x)^+=max(0,x)=ReLu(x)$. In practice, we often do segmented trainings and collect parameters as well as gradient values associated with training date in each segment. If the sampling interval is $\Delta  t$, training period $T=l\Delta t$, $s_k=s_{k\Delta t}$, $a_k=a_{k\Delta t}$, then the regularizers could be rewritten as :
$$J_{Lip}(\theta_p)\!=\!\lambda_1\sum_{k=0}^{l-1}(\Vert\partial f^{\theta_p}(s,a)\Vert_2\!+\!\Vert\partial g^{\theta_p}(s,a)\Vert_2\!-\!D_1)^+\eqno(8)$$
$$J_{Lip}(\theta_v)\!=\!\lambda_2\sum_{k=0}^{l-1}(\Vert\partial \mu^{\theta_v}(s,a)\Vert_2\!+\!\Vert\partial \sigma^{\theta_v}(s,a)\Vert_2\!-\!D_2)^+\eqno(9)$$\par
These two regulation terms are added in the objective function of $\theta_p$ and $\theta_v$ respectively.
\subsection{Value Estimator(VE)}
We introduce the value estimator VE first since it is the core of reinforcement learning. Agents base on the Q function in VE to form expectations on future reward under current environment state and its current action policy. Note that the classical, discretized Q function under optimal parameters should satisfy:
$$Q^{\theta^*}(s_k,a_k)=E^{(s_k,a_k)}[R_k+\gamma Q^{\theta^*}(s_{k+1},a_{k+1})]\eqno(10)$$
where $E^{(s_k,a_k)}[\cdot]$ is the conditional expectation, which is the expectation under known conditions $(s_k,a_k)$, and $\theta^*$ is the optimal parameters. It is obvious that random varible $Q^{\theta^*}(s_k,a_k)$ satisifies Markov property in discrete time. $Q^{\theta^*}(s,a)$ is therefore recursively determined by
\begin{flalign*}
Q^{\theta^*}(s,a)&=Q^{\theta^*}(s_0,a_0)\\
&=E^{(s,a)}[R_0+\gamma Q^{\theta^*}(s_1,a_1)]\\
&=E^{(s,a)}[R_0+\gamma E^{(s_1,a_1)}[R_1+\gamma Q^{\theta^*}(s_2,a_2)]]\\
&=E^{(s,a)}[R_0+\gamma E^{(s,a)}[R_1+\gamma Q^{\theta^*}(s_2,a_2)|\mathcal{N}_1]]\\
&=E^{(s,a)}[R_0+\gamma R_1+\gamma^2 Q^{\theta^*}(s_2,a_2)]
\end{flalign*}
where $\mathcal{N}_k$ represents the $\sigma$ algebra which is defined by $\left\{(s_l,a_l)\right\}_{l\leq k}$; $R_k$ is the reward at the $k$ moment. So by recursive substitution, it is easy to find that at $k$ moment, 
$$Q^{\theta^*}(s,a)=E^{(s,a)}\left[\sum_{i=0}^{k-1}{\gamma^i R_i}+\gamma^k Q^{\theta^*}(s_k,a_k)\right]\eqno(11)$$\par
This constraint for Q function in discrete cases can be easily extended to the continuous case. If the time interval is $\Delta t$, the reward in the $k\Delta t$ moment is $R_k$, the discrete time $k$ can be replaced by the continuous time $t$:
$$Q^{\theta^*}(s,a)=E^{(s,a)}\left[\int_{0}^{t}{\gamma^s r_sdt}+\gamma^t Q^{\theta^*}(s_t,a_t)\right]\eqno(12)$$
where $r_s$ is the rate of the change of the reward. This equation is the constraint for Q function of VE in IRL. Now it raises two questions for Q function:
\begin{itemize}
\item Does a Q function that satisfies the constraint equation (12) exist?
\item If such a Q function exists, how do we update the parameters of the network to approximate the optimal parameters and simultaneously satisfy the constraint equation(12) during the training process?
\end{itemize}\par
Before answering these two questions, it is necessary to make some changes to the constraint equation (12). Let take the derivative of Q function with respect to t and make the first order condition equal to 0:
$$0\!=\!\frac{\partial}{\partial t}Q^{\theta^*}\!(s,a)\!=\!\frac{\partial}{\partial t}E^{(s,a)}\!\left[\int_{0}^{t}\!{{\gamma^s r_sdt}\!+\!\gamma^t Q^{\theta^*}\!(s_t,\!a_t)}\right]\eqno(13)$$
Since the actual process $\int_{0}^{t}{{\gamma^s r_sdt}+\gamma^t Q^{\theta^*}(s_t,a_t)}$ is always bounded, dominated convergence theorem allows us to substitute the partial operator with expectation operator, so 
\begin{flalign*}
0&=E^{(s,a)}\left[\frac{\partial}{\partial t}\left(\int_{0}^{t}{{\gamma^s r_sdt}+\gamma^t Q^{\theta^*}(s_t,a_t)}\right)\right]\\
&=E^{(s,a)}[\gamma^t r_t+ln\gamma\cdot\gamma^t Q^{\theta^*}(s_t,a_t)+\gamma^t\frac{\partial}{\partial t}Q^{\theta^*}(s_t,a_t)]\\
&=E^{(s,a)}[r_t+ln\gamma\cdot Q^{\theta^*}(s_t,a_t)+\frac{\partial}{\partial t}Q^{\theta^*}(s_t,a_t)]
\end{flalign*}\par
By setting $y=(s,a), Y_t=(s_t,a_t)$ and $q(y,t)=E^{y}[Q^{\theta^*}(Y_t)]$, the following formula is obtained:
$$E^{(s,a)}[r_t]+ln\gamma\cdot q(y,t)+\frac{\partial}{\partial t}q(y,t)=0\eqno(14)$$
The question is how to deal with the partial derivative of $q(y,t)$, or $E^{(s,a)}[Q^{\theta^*}(s_t,a_t)]$ with respect to time. Fortunately, it can be solved by using the theory of stochastic differential equations. In the following we show the definition of the characteristic operator of stochastic differential equations.

{\bf Definition}. {Characteristic Operators of I\^{t}o Diffusion and Their Expressions}
\mbox{}\par
\rm Suppose $\{X_t\}$ is a I\^{t}o diffusion process on $\mathbb{R}^n$,  the generator of $X_t$ can be defined as:
$$Af(x)=\lim\limits_{t\downarrow0}{\frac{E^x[f(X_t)]-f(x)}{t}},x\in\mathbb{R}^n\eqno(15)$$
So, the set of all functions $f$ that have the upper limit at $x$ can be defined as $\mathcal{D}_A(x)$ where $\mathcal{D}_A$ is the set of functions whose upper limits exist when $f\in C^2,f\in \mathcal{D}_A$. $\mathcal{D}_\mathcal{A}$ is also called characteristic operator. These two operators are equivalent when $f\in C^2$.\par
Suppose $X_t^x$ is a I\^{t}o diffusion process with an initial condition $X_0^x=x$ on $\mathbb{R}^n$, so
$$X_t^x(\omega)=x+\int_{0}^{t}{u(s,\omega)ds}+\int_{0}^{t}{v(s,\omega)\cdot dB_s(\omega)}\eqno(16)$$\par
Suppose $f\in C^2_0(\mathbb{R}^n)$, $\tau$ is a stopping time related to $\mathcal{F}_t^{(n)}$ with the condition $E^x[\tau]<\infty$, and $X_t^x$ is bounded on the support set of f, then 
\setcounter{equation}{16}
\begin{equation}
\begin{aligned}
E^x[f(X_\tau)]\!=\!f(x)\!+\!E^x\!\left[\!\int_{0}^{\tau}\!\left(\!\sum_{i}\!{u_i(s,\omega) \frac{\partial f}{\partial x_i}(Y_s)}\right)ds\right. \\
 \left.\!+\!\frac{1}{2}\!\int_{0}^{\tau}\!\left(\!\sum_{i,j}\!(vv^T)_{ij}(s,\omega) \frac{\partial^2 f}{\partial x_i\partial x_j}(Y_s) \!\right)\!ds\!\right]
\end{aligned}
\end{equation}
and the characteristic operator of I\^{t}o diffusion process $\mathcal{A}$ can be defined as:
$$\mathcal{A}=\sum_{i}{u_i(x)\frac{\partial}{\partial x_i}} +\frac{1}{2}\sum_{i,j}(vv^T)_{ij}(x) \frac{\partial^2 }{\partial x_i\partial x_j}\eqno(18)$$
which is a second order linear partial differential operator.\par
The next lemma shows the relationship between $\frac{\partial}{\partial t}E^x[f(X_t)]$ and the characteristic operator.
\begin{lemma}{Kolmogorov backward equation} 
\label{rem: Kolmogorov backward equation}
	\mbox{}\par
	\rm Suppose $f\in C_0^2(\mathbb{R}^n)$, define 
	$$u(t,x)=E^x[f(x)]\eqno(19)$$
	$(i)$ If for every t, $u(t,\cdot)\in \mathcal{D}_\mathcal{A}$, then
	$$\frac{\partial u}{\partial t}=\mathcal{A}u,t>0,x\in \mathbb{R}^n\eqno(20)$$
	$$u(0,x)=f(x),x\in \mathbb{R}^n\eqno(21)$$
	$(ii)$ If there exists a $w(t,x)\in C^{1,2}(\mathbb{R}\times \mathbb{R}^n)$ that satisfies two conditions above, we have 
	$$w(t,x)=u(t,x)=E^x[f(X_t)].\eqno(22)$$
\end{lemma}\par
Through applying the lemma \ref{rem: Kolmogorov backward equation}, Kolmogorov backward equation, to Q function, it is possible to calculate $\frac{\partial}{\partial t}E^y[Q^{\theta^*}(Y_t)]$. From the diffusion equation of $Y_t$, the characteristic operator $\mathcal{A}_Y$ of $Y_t$ can be written as
$$\mathcal{A}_Y=\sum_{i=1}^{m+n}{F_i^{(\theta_p,\theta_v)}\frac{\partial }{\partial y_i}}+\frac{1}{2}\sum_{i,j=1}^{m+n}{(GG^T)_{ij}^{(\theta_p,\theta_v)}\frac{\partial^2 }{\partial y_i\partial y_j}}\eqno(23)$$
or
$$\mathcal{A}_Y=F^{(\theta_p,\theta_v)}\cdot\partial+\frac{1}{2}(GG^T)^{(\theta_p,\theta_v)}*\partial^2 \eqno(24)$$
where $*$ is the operator that $A*B=\sum_{ij}{A_{ij}\cdot B_{ij}}$. Let consider 
$$q(y,t)=E^y[Q^{\theta^*}(Y_t)]=E^{(s,a)}[Q^{\theta^*}(s_t,a_t)]\eqno(25)$$
According to Kolmogorov backward equation, it is obvious that
\begin{small}
\begin{flalign*}
\frac{\partial}{\partial t}q(y,\!t)&\!=\mathcal{A}_Yq(y,t)\\
&=\sum_{i=1}^{m+n}{F_i^{(\theta_p,\theta_v)}(y)\frac{\partial q}{\partial y_i}(y,t)}\\
&+\frac{1}{2}\sum_{i,j=1}^{m+n}{(GG^T)_{ij}^{(\theta_p,\theta_v)}(y)\frac{\partial^2 q}{\partial y_i\partial y_j}(y,t)}\\
&\!=\!F^{(\theta_p,\theta_v\!)}(y)\!\cdot\!\partial_y q(y,\!t)\!+\!\frac{1}{2}(GG^T\!)^{(\theta_p,\theta_v)}(y)\!*\!\partial^2_y q(y,\!t)\\
\end{flalign*}
\end{small}
Combined with eq(14), we get the constraint equation of $q(y,t)$:
\setcounter{equation}{25}
\begin{equation}
\begin{aligned}
E^{(s,a)}[r_t]&+ln\gamma\cdot q(y,t)+\sum_{i=1}^{m+n}{F_i^{(\theta_p,\theta_v)}(y)\frac{\partial q}{\partial y_i}(y,t)}\\
&+\frac{1}{2}\sum_{i,j=1}^{m+n}{(GG^T)_{ij}^{(\theta_p,\theta_v)}(y)\frac{\partial^2 q}{\partial y_i\partial y_j}(y,t)}=0
\end{aligned}
\end{equation}
or
\begin{equation}
\begin{aligned}
E^{(s,a)}[r_t]&+ln\gamma\cdot q(y,t)+F^{(\theta_p,\theta_v)}(y)\cdot\partial_y q(y,t)\\
&+\frac{1}{2}(GG^T)^{(\theta_p,\theta_v)}(y)*\partial^2_y q(y,t)=0
\end{aligned}
\end{equation}\par
A new second order elliptic partial differential operator appears as following
$$L_Y\!=\!ln\gamma+\mathcal{A}_Y\!=\!ln\gamma\!+\!F^{(\theta_p,\theta_v)}\!\cdot\!\partial_y \!+\!\frac{1}{2}(GG^T)^{(\theta_p,\theta_v)}\!*\!\partial^2_y\eqno(28)$$
And eq(27) can be rewritten as
$$L_Yq(y,t)=-E^{y}[r_t]\eqno(29)$$
This formula indicates that if Q function exists, $q(y,t)$ could be the solution of a class of second order elliptic partial differential equations. In other words, IRL is related to partial differential equation theory.\par
The next step is eliminating the influence of the time, we discuss the case for $t=0$. By setting $$q(y)=q(y,0)=Q^{\theta^*}(y)=Q^{\theta^*}(s,a)\eqno(30)$$
from the initial formula eq(12) we get
$$q(y)=E^{y}\left[\int_{0}^{t}{\gamma^s r_sdt}\right]+\gamma^tq(y,t)\eqno(31)$$
Then apply operator $L_Y$ to both sides of the equation:
\begin{flalign*}
L_Yq(y)&=L_YE^{y}\left[\int_{0}^{t}{\gamma^s R_sds}\right]+\gamma^tL_Yq(y,t)\\
&=L_YE^{y}\left[\int_{0}^{t}{\gamma^s R_sds}\right]-\gamma^tE^{y}[r_t]
\end{flalign*}
The left side of the equation doesn't contain time variable, so 
$$\frac{\partial}{\partial t}\left\{L_YE^{y}\left[\int_{0}^{t}{\gamma^s R_sdt}\right]-\gamma^tE^{y}[r_t]\right\}=0\eqno(32)$$
which means the right side of the equation only relate to $y$, written as $\Phi (y)$(that is also the $-E^y[r_0]=-E^y[r]$). Now we get a partial differential equation of $q(y)$, $Q^{\theta^*}(s,a)$ and such a Q function dose exist according to the partial differential equation theory. (Actually it is a Dirichlet-Poisson question.) However, if we want to use PDE theory, it is necessary to discuss the boundary condition of the Q function. Fortunately in practice, agent's action policies are generally bounded. It is hard to know the exact boundary conditions of environment states, but they are bounded in a large enough area. We write the domain of $Q^{\theta^*}(s,a)$ as $I(s,a)=I(y)$, and 
$$q(y)=0\ on\ \mathbb{R}^{m+n}-I(y)\ and\ \partial I(y)\eqno(33)$$
where $\partial I(y)$ is the boundary of $I(y)$.\par
Now we have two equations:
$$L_Yq(y)=\Phi(y), y\in I(y)\eqno(34)$$
$$q(y)=0, y\in \partial I(y)\eqno(35)$$
where $L_Y$ is a second-order elliptic partial differential operator. This is a typical Dirichlet question. Here we only demonstrate the main theorem of the existence and uniqueness theorem of the solution to Dirichlet question. Detailed proof can be found in \cite{DG:98}.

\begin{lemma}{Existence and uniqueness of the solution to  Dirichlet question}
\label{rem: Existence and uniqueness of the solution of  Dirichlet question}
	\mbox{}\par
	\rm Suppose
	 $$L=\sum_{i,j}{a_{ij}(x)}\frac{\partial^2}{\partial x_i\partial x_j}+\sum_{i}{b_{i}(x)}\frac{\partial}{\partial x_i}+c(x)$$ 
	 is strictly elliptical in a bounded area $\Omega$, and there exists a positive number $\lambda$ such that
	 $$\sum_{i,j}{a_{ij}(x)\xi_i\xi_j}\geq\lambda\Vert\xi\Vert_2^2,\forall x\in\Omega,\xi\in\mathbb{R}^n$$
	 of $c(x)<0$. If $f$ and the coefficient function of $L$ belong to $C^\alpha(\overline{\Omega})$, (which means they have $\alpha$-order H\"{o}lder continuity) $\Omega$ is a $C^{2,\alpha}$ area and $\phi\in C^{2,\alpha}(\overline{\Omega})$, then Dirichlet question 
	 $$Lu=f,\ in\ \Omega\quad; u=\phi\ on\ \partial\Omega$$
	 has a unique solution that belongs to $C^{2,\alpha}(\overline{\Omega})$.
\end{lemma}\par
In practice, when $ln\gamma<0$, we might consider $\Phi(y)$ is at least second order continuous and $\alpha$-order H\"{o}lder continuous. This gives the existence and uniqueness of the Q-function. However, if Q function exists, it must be at least second order continuous, which means any nonlinear functions that are not second order continuous (such as ReLU) cannot be used in the nonlinear function layers in the networks. On the other hand, sigmoid, which is both bounded and $C^\infty$ continuous, is apparently a better choice.\par
Once Q function exists, we can use optimization methods such as gradient descent, SGD, Adam algorithm to optimize
$$J_Q(\theta)=(L_YQ^{\theta}(s,a)+E^{y}[r])^2,\eqno(36)$$
 and derive the optimal parameters $\theta^*$. In details,
\setcounter{equation}{36}
\begin{equation}
\begin{aligned}
J_Q(\theta)&\!=\!\!\left(\!ln\gamma\cdot Q^{\theta}(s,a)\!+\!\!\sum_{i=1}^{m+n}{F_i^{(\theta_p,\theta_v)}(s,a)\frac{\partial Q^{\theta}}{\partial y_i}(s,a)}\right. \\
&\left.+\frac{1}{2}\!\!\sum_{i,j=1}^{m+n}\!\!(GG^T\!)_{ij}^{(\theta_p,\!\theta_v\!)}(s,a)\frac{\partial^2 Q^{\theta}}{\partial y_i\partial y_j}(s,\!a)\!+\!\!E^{y}[r]\!\right)^2.
\end{aligned}
\end{equation}
This is the objective function used to update the network. Because it satisfies the Markov property, we can also use the training data in each step $(s_k,a_k)$ of the process as the initial value. Suppose the number of training samples in one step is $l$, then:
\setcounter{equation}{37}
\begin{equation}
\begin{split}
J_Q(\theta)=&\sum_{k=0}^{l}\left(
E^{(s_{k},\!a_{k})}[r_k]\!+ln\gamma\!\cdot\! Q^{\theta}(s_k,a_k)\right.\\
&\left.+\frac{1}{2}\sum_{i,j=1}^{m+n}\!(GG^T)_{ij}^{(\theta_p,\theta_v)}(s_k,\!a_k)\frac{\partial^2 Q^{\theta}}{\partial y_i\partial y_j}(s_k,\!a_k)\!\right.\\
&\left.+\!\sum_{i=1}^{m+n}\!{F_i^{(\theta_p,\theta_v)}(s_k,\!a_k)\frac{\partial Q^{\theta}}{\partial y_i}(s_k,\!a_k)}\right)^2
\end{split}
\end{equation}
\par
In practice, it is difficult to solve such a partial differential equation with uncertain parameters by training the network through gradient descent, and the gradient can easily get too large and impede training preceeding. This can be alleviated by adding an additional objective function. Recall the Q function in equation (12).
During the training process, we record the next state $(s_k',a_k')$ and the above Q function can be written in discrete time:
$$Q^{\theta^*}(s_k,a_k)\approx E^{(s_k,a_k)}\left[r_k\Delta t\right]+\gamma^{\Delta t} Q^{\theta^*}(s_k',a_k')\eqno(40)$$ 
Then we get an additional objective function:
$$J_Q'(\theta)=\sum_{k=0}^{l}(Q^{\theta}(s_k,a_k)-r_k\Delta t-\gamma^{\Delta t} Q^{\theta}(s_k',a_k'))^2\eqno(41)$$ \par
Since the integration is discretizated, this objective function is less precise than the one with second-order gradient terms that we obtained earlier. However, this objective function is more intuitive and its convergence speed can be accelerated in training.

\subsection{Environment State Estimator(ESE)}
ESE is not necessary in classical reinforcement learning algorithms such as A3C and DDPG. However in IRL, ESE is essential to make $(s_t,a_t)$ I\^{t}o diffusion.  Note that there is a fundamental difference bewteen $\theta_p$ and $(\theta,\theta_v)$: $\theta_p$ comes from the state change of the environment itself, so this parameter is objective; $(\theta,\theta_v)$ are related to rewards set by human, so they are subjective. The difference makes ESE network updating process in IRL a process of parameter estimation.
Now we raise two theorems as the starting point of the ESE update process.

\begin{lemma}{Girsanov Theorem and Likelihood Function}
\label{rem: Girsanov Theorem and Likelihood Function}
	\mbox{}\par
	\rm Suppose
	$$Y(t)=\zeta(t,\omega)dt+\zeta(t,\omega)\cdot dB_t,$$
where $t<=T,\ \beta(t,\omega)\in\mathbb{R}^n,\ \zeta(t,\omega)\in\mathbb{R}^{n\times m}$. Suppose there exist processes $u(t,\omega)\in\mathcal{W}^m_\mathcal{H},\alpha(t,\omega)\in\mathcal{W}^m_\mathcal{H}$ such that
	$$\zeta(t,\omega)u(t,\omega)=\beta(t,\omega)-\alpha(t,\omega)$$
and assume that $u(t,\omega)$ satisfies Novikov's condition
$$E \left[exp \left(\frac{1}{2}\int_0^T{u^2(s,\omega)ds}\right)\right]<\infty.$$\par
If
	$$M_t\!=\!exp\left(\!\!-\!\!\int_0^t\!{u(s,\omega)\cdot dB_s}\!-\!\frac{1}{2} \int_0^t\!{\Vert u(s,\omega)\Vert_2^2ds}\!\right),\ t\!<=\!T$$
and
$$dQ(\omega)=M_T(\omega)dP(\omega)\ \ \ \ on\  \mathcal{F}_T^{(m)},$$
then
$$\hat{B}_t:=\int^t_0{u(s,\omega)ds}+B(t)$$
is a Brownian motion w.r.t. Q and the process $Y\left(t \right)$ has the stochastic integral representation
	$$Y(t)=\alpha(t,\omega)dt+\zeta(t,\omega)d\hat{B}_t.$$
\end{lemma}~\\ \par
Consider the I\^{t}o difussion $dX_t=a^{\theta}(X_t)dt+b^{\theta}(X_t)\cdot dB_t$, applying Girsanov Theorem conditional on $b$ is reversible and $\alpha=0$, then
	$$u^{\theta}(X_t)=u(X_t,\theta)=(b^{-1})^{\theta}(X_t)\cdot a^{\theta}(X_t)$$
	and the log-likelihood function is defined as:
	$$l_T(\theta)=ln\frac{dQ(\omega)}{dP(\omega)}=lnM^{\theta}_T(\omega)$$
	that is 
	$$l_T(\theta)=-\int_0^t{u(s,\omega)\cdot dB_s}-\frac{1}{2} \int_0^t{\Vert u(s,\omega)\Vert_2^2ds}\eqno(42)$$
Recall that in IRL,   the difussion process $Y_t=(s_t,a_t)$ is 
$$dY_t =F^{(\theta_p,\theta_v)}(Y_t)dt+G^{(\theta_p,\theta_v)}(Y_t)\cdot d\overline{B}_t,Y_t\in \mathbb{R}^{m+n}.$$
According to lemma \ref{rem: Girsanov Theorem and Likelihood Function}, 
$$u(Y_t,\theta_p)=u^{\theta_p}(Y_t)=(G^{-1})^{(\theta_p,\theta_v)}(Y_t)\cdot F^{(\theta_p,\theta_v)}(Y_t)\eqno(43)$$
where $u(Y_t,\theta_p)\in\mathbb{R}^{n+m} $, $\theta_v$ is a constant parameter set that does not engage in ESE parameters update process. Then, the log-likelihood function $l_T(\theta_p)$ can be defined as:
$$l_T(\theta_p)=-\int_0^t{u(Y_s,\theta_p)\cdot dB_s}-\frac{1}{2} \int_0^t{\Vert u(Y_s,\theta_p)\Vert_2^2ds}\eqno(44)$$
The next lemma shows how to use this log-likelihood function to estimate parameter $\theta_p$. The detail of the proof can be found in\cite{WC:16}.

\begin{lemma}{Likelihood Functions for $\theta_p$ Estimation}
\label{rem: 5}
	\mbox{}\par
	\rm Suppose $X_t\in\mathbb{R}^n$ is a stationary and ergodic process, and
	$$dX_t=a^{\theta}(X_t)dt+b^{\theta}(X_t)\cdot dB_t$$
	$$u^{\theta}(x)=u(x,\theta)=(b^{-1})^{\theta}(x)\cdot a^{\theta}(x)$$
	$P_{\theta}$ is a probability measure obtained in the Girsanov theorem. $l_T(\theta)$ is the log-likelihood function. If the conditions below hold:\\
	(A1) $P_\theta\ne P_{\theta'}$, while $\theta$ is the true value and $\theta\ne {\theta}'$;\\
	(A2) $P_\theta\left(\left\{\int_0^T{\Vert u(X_s,\theta)\Vert_2^2ds}<\infty \right\}\right)=1,\forall T>0$, while $\theta$ is the true value;\\
	(A3) Suppose $I(\theta)$ is the domain of $\theta$ of $u(x,\theta)$, then for $r\in I(\theta)$,
	$$|u(x,r)|\leq M(x),E_{\theta}[M(X_0)]^2<\infty$$
    $$|\partial _su(x,r)|\leq Q(x),E_{\theta}[Q(X_0)]^2<\infty$$
	where $E_\theta$ is the expectation based on $P_\theta$; $|\cdot|$ is some norm;\\
	(A4) $\forall r\in I(\theta),E_{\theta}\left[\int_0^T{\Vert u(X_s,r)\Vert_2^2ds}\right]<\infty$.\\
	Then\\
	(1) The solution $\hat{\theta}_T$ of equation $l_T'(\theta)=0$ satisfies $$\hat{\theta}_T\xrightarrow{a.s.}\theta\ when\ T\to\infty\eqno(45)$$
	(2) When $T\to\infty$, $$\sqrt{T}(\hat{\theta}_T-\theta)\xrightarrow{a.s.}N\left(0,\left(E_{\theta}\left[\partial_{\theta}u(X_0,r)\right]\right)^{-1}\right)\eqno(46)$$	
\end{lemma}\par

This theorem gives the error distribution of $\sqrt{T}(\hat{\theta}_T-\theta)$ when T approaches infinity. The ergodicity of stationary states the time average of a stochastic process (the mean of the integral of time) equivalent to the space average (expectation),
$$\frac{1}{T}\int_{0}^{T}{f(X_t)dt}\xrightarrow{a.s.}E[f(\xi)] $$
$f$ is a measurable function which satisfies $E[f(\xi)]<\infty$, where $\xi$ is a random variable under a unknown distribution. Generally, if $X_t$ is a solution to a certain stochastic differential equation, the ergodicity can be satisfied by adding constraints to the coefficients of the equation. If there are special requirements, regularizer can be added to the network to force it to meet the requirements. But in practice, it is not necessary because our approximation is always limited in time. We only need to ensure the corresponding accuracy. Condition A1 is always satisfied, and for other conditions, if we can guarantee that $u(y,\theta)\in C^2_0$, they will be met automatically. Since the input and output of the network are always bounded, we only need to ensure that the non-linear layers of the network structure are at least second order continuous. Unfortunately, ReLU was once again disqualified under these conditions, while some $C^\infty$ nonlinear function such as sigmoid meet all requirements. From lemma \ref{rem: 5}, we can get the objective function of ESE, that is, the discretization of eq(45) and (46).
\setcounter{equation}{46}
\begin{equation}
\begin{aligned}
u((s_k,a_k),\theta_p)&\!=\!u^{\theta_p}(s_k,a_k)&\\
&\!=\!\!(G^{-1}\!)^{(\theta_p,\theta_v\!)}(s_k,\!a_k)\!\cdot\! F^{(\theta_p,\theta_v\!)}(s_k,\!a_k)
\end{aligned}
\end{equation}
$$l_T(\theta_p)\!=\!\!-\!\!\sum_{k=0}^{1}{u^{\theta_p}(s_k,a_k)\!\cdot \!\Delta \overline{B}_k}\!-\!\frac{\Delta t}{2}\!\sum_{k=0}^{l}\!{\Vert u^{\theta_p}(s_k,a_k)\Vert_2^2}\eqno(48)$$
where $\Delta B_k=B_{k+1}-B_k$ (which is a main property of I\^{t}o integral). From the properties of Brownian motion we can get $\Delta B_k\sim N(0,\Delta t)$. The random variables are eliminated after taking expected values. Then the objective function of ESE becomes:
$$J_E(\theta_p)=\frac{\Delta t}{2}\sum_{k=0}^{l}{\Vert u^{\theta_p}(s_k,a_k)\Vert_2^2}$$
\par
Because only $\theta_p$ is contained in the objective function, it is apparent that it is possible to eliminate some items which only contain $\theta_v$ to reduce computation load. Since 
$$ 
F^{(\theta_p,\theta_v)}(s_k,a_k)=
\left(\begin{array}{c}
f^{\theta_p}(s_k,a_k)\\
\mu^{\theta_v}(s_k,a_k)
\end{array}\right)
$$
$$ 
G^{(\theta_p,\theta_v)}(s_k,a_k)=
\left(\begin{array}{cc}
g^{\theta_p}(s_k,a_k)&0\\
0&\sigma^{\theta_v}(s_k,a_k)
\end{array}\right)
$$
the $u^{\theta_p}(s_k,a_k)$ has the form:
$$
u^{\theta_p}(s_k,a_k)=
\left(\begin{array}{c}
(g^{-1})^{\theta_p}(s_k,a_k)\cdot f^{\theta_p}(s_k,a_k)\\
(\sigma^{-1})^{\theta_v}(s_k,a_k)\cdot\mu^{\theta_v}(s_k,a_k)
\end{array}\right)
\eqno(49)$$
which means $\theta_v$ can be eliminated during the update process of $\theta_p$. Finally, 
$$u^{\theta_p}(s_k,a_k)=(g^{-1})^{\theta_p}(s_k,a_k)\cdot f^{\theta_p}(s_k,a_k)\eqno(50)$$
$$J_E(\theta_p)=\frac{\Delta t}{2}\sum_{k=0}^{l}{\Vert u^{\theta_p}(s_k,a_k)\Vert_2^2}\eqno(51)$$
In addition, we can control the accuracy of the error. From the second conclusion of lemma 5, we can see that if the time period is long enough,
$$\sqrt{l\Delta t}(\theta_p-\theta_p^*)\sim N\left(0,\left(E_{\theta}\left[\partial_{\theta}u^{\theta_p}(s_0,a_0)\right]\right)^{-1}\right)\eqno(52)$$
where $\theta_p^*$ is the truth value. Generally, $(s_0,a_0)$ in the control process can be approximated by fixed endpoints $(s,a)$, then
$$\sqrt{l\Delta t}(\theta_p-\theta_p^*)\sim N\left(0,\left(\partial_{\theta}u^{\theta_p}(s,a)\right)^{-1}\right).\eqno(53)$$
By setting a reasonable initial value and a long enough runtime, the error of the estimated parameters can be properly controlled.\par

On the other hand, similar to previous value estimators, the environmental state estimator introduces an additional objective function to estimate the environmental state. Similar to our previous discussion, we record the next state $(s_k',a_k')$ so the additional objective function of ESE is:
$$J_E'(\theta_p)\!=\!\!\sum_{k=0}^{l}\!E\left[\Vert s_k\!+\!f^{\theta_p}(s_k,a_k)\Delta t\!+\!g^{\theta_p}(s_k,a_k)\cdot B_t\!-\!s_k'\Vert_2^2\right]$$
The simplified formula is as follows:
\begin{equation}
\begin{aligned}
J_E'(\theta_p)=&\sum_{k=0}^{l}\left[(s_k+f^{\theta_p}(s_k,a_k)\Delta t-s_k')^2\right.\\
&\left.+ \Delta t\cdot det\left((gg^T)^{\theta_p}(s_k,a_k)\right)\right]
\end{aligned}
\end{equation}
Generally, the variance term $g$ can be neglected to simplify the calculation, so
$$J_E'(\theta_p)=\sum_{k=0}^{l}\left[(s_k+f^{\theta_p}(s_k,a_k)\Delta t-s_k')^2\right].$$
This additional objective function of ESE can be used to measure the accuracy of ESE in estimating future state changes.

\subsection{Action Policy Generator (APG)}
The policy gradient method can be used to update the network parameters in the action policy generator (APG). In classical reinforcement learning, given the objective function of action policy network $J_A(\theta_v)$, we have
$$\partial J_A(\theta_v)=E\left[\partial_{\theta_v}ln\pi^{\theta_v}(a_k|s_k)\cdot Q^{\theta}(s_k,a_k)\right]\eqno(54)$$
where $\pi^{\theta_v}(a_k|s_k)$ is the conditional distribution of action $a_k$ when current environment state $s_k$ is known. This policy gradient updating method can be used in IRL after making some modifications.\par
Recall that
$$da_t =\mu^{\theta_v}(s_t,a_t)dt+\sigma^{\theta_v}(s_t,a_t)\cdot d\tilde{B}_t.$$\par
Suppose $\Delta t$ is the minimal time interval, so from the above SDE, we can see that in a short period of time
$$\Delta a_k\approx \mu^{\theta_v}(s_k,a_k)\Delta t+\sigma^{\theta_v}(s_k,a_k)\cdot \Delta\tilde{B}_t\eqno(55)$$
or
$$a_{k+1}\approx a_k+ \mu^{\theta_v}(s_k,a_k)\Delta t+\sigma^{\theta_v}(s_k,a_k)\cdot \Delta\tilde{B}_k\eqno(56)$$
where 
$$\Delta \tilde{B}_k=\tilde{B}_{k+1}-\tilde{B}_k\sim N(0,\Delta t),$$
so action $a_{k+1}$ follows the conditional normal distribution with the condition $(s_k,a_k)$, or 
$$a_{k+1}\!\sim\! N\left(a_k\!+\!\mu^{\theta_v}(s_k,a_k)\Delta t,\left((\sigma^T\!\sigma)\right)^{\theta_v}\!(s_k,a_k)\Delta t\right)\eqno(57)$$\par 
Then we get the expression of conditional probability distribution of $a_{k+1}$ as follows, 
\setcounter{equation}{57}
\begin{equation}
\begin{aligned}
\pi^{\theta_v}&(a_{k+1}|s_k,a_k)\\
=&\frac{{|det(\sigma^{\theta_v}(s_k,a_k))|}^{-1}}{(2\pi)^{\frac{m}{2}}\sqrt{\Delta t}}exp\left(-\frac{1}{2\Delta t}\Vert{(\sigma^{-1})}^{\theta_v}\right.\\
&\left.\cdot(s_k,a_k)(a_{k+1}-a_k-\mu^{\theta_v}(s_k,a_k)\Delta t)\Vert_2^2\right)
\end{aligned}
\end{equation}
take logarithms on both sides,
\setcounter{equation}{58}
\begin{equation}
\begin{split}
ln\pi^{\theta_v}(a_{k+1}|s_k,a_k)=&-ln|det(\sigma^{\theta_v}(s_k,a_k))|-\frac{1}{2\Delta t}\\
&\cdot\Vert{(\sigma^{-1})}^{\theta_v}(s_k,a_k)(a_{k+1}-a_k\\
&-\mu^{\theta_v}(s_k,a_k)\Delta t)\Vert_2^2+C
\end{split}
\end{equation}
where $C$ is a constant independent of $\theta_v$, so $\partial J_A(\theta_v)$ can be written as
\setcounter{equation}{59}
\begin{equation}
\begin{split}
\partial J_A(\theta_v\!)\!\approx& \partial_{\theta_v}\!\bigg\{\!ln|(\sigma^{\theta_v}(s_k,\!a_k))|\!+\!\frac{1}{2\Delta t}\Vert(\sigma^{-1}\!)^{\theta_v}(s_k,\!a_k)\\
&\cdot(a_{k+1}\!-\!a_k\!-\!\mu^{\theta_v}(s_k,\!a_k)\Delta t)\Vert_2^2\!\bigg\}\!\cdot\! Q^{\theta}(s_{k},\!a_{k})
\end{split}
\end{equation}\par
After discretization and sampling data $(s_k,a_k)$ for several time periods, we have 
\begin{equation}
\begin{split}
&\partial J_A(\theta_v)\!\approx \!\sum_{k=0}^{l}\!\partial_{\theta_v}\!\bigg\{\!ln|det(\sigma^{\theta_v}(s_k,\!a_k))|\!+\!\frac{1}{2\Delta t}\Vert(\sigma^{-1}\!)^{\theta_v}\\
&\cdot (s_k,\!a_k)(a_{k+1}\!-\!a_k\!-\!\mu^{\theta_v}(s_k,\!a_k)\Delta t)\Vert_2^2\bigg\}\!\cdot\! Q^{\theta}(s_{k},\!a_{k})
\end{split}
\end{equation}
Variance can be ignored in the update period, so
\begin{equation}
\begin{split}
J_A(\theta_v)\approx& \sum_{k=0}^{l}\frac{1}{2\Delta t}Q^{\theta}(s_{k},a_{k})\Vert(\sigma^{-1})^{\theta_v}(s_k,a_k)\\
&\cdot(a_{k+1}-a_k-\mu^{\theta_v}(s_k,a_k)\Delta t)\Vert_2^2
\end{split}
\end{equation}
which is one objective function of APG.\par
From the definition of the Q function, another objective function of action strategy generator is obtained. Obviously, we hope that the action policy made by the agent maximize the Q value for the next moment. This means that agent's action policy should maximize the gradient of Q function.
$$\lim_{t\to 0}\frac{\partial}{\partial t}E^{(s,a)}\left[Q^{\theta}(s_t,a_t)\right]\eqno(63)$$\par
We have learned that according to Kolmogorov's backward equation,
\setcounter{equation}{63}
\begin{equation}
\begin{split}
\lim_{t\to 0}\frac{\partial}{\partial t}E^{(s,a)}\!\left[Q^{\theta}(s_t,a_t)\right]\!&=\!\lim_{t\to 0}\mathcal{A}_Y E^{(s,a)}\!\left[Q^{\theta}(s_t,a_t)\right]\!\\
&\!=\!\mathcal{A}_YQ^{\theta}(s,a)
\end{split}
\end{equation}
where
$$\mathcal{A}_Y=\sum_{i=1}^{m+n}{F_i^{(\theta_p,\theta_v)}\frac{\partial }{\partial y_i}}+\frac{1}{2}\sum_{i,j=1}^{m+n}{(GG^T)_{ij}^{(\theta_p,\theta_v)}\frac{\partial^2 }{\partial y_i\partial y_j}}\eqno(65)$$
By maximizing the parameters $\theta_v$ in $\mathcal{A}_YQ^{\theta}(s,a)$ we ensure that the action policy generated by the agent makes the Q value increasing continuously. Finally, we have
\setcounter{equation}{65}
\begin{equation}
\begin{split} 
J'_A(\theta_v)=&-\sum_{k=0}^{l}\left(\sum_{i=1}^{m+n}{F_i^{(\theta_p,\theta_v)}\frac{\partial Q^{\theta}}{\partial y_i}}(s_k,a_k)\right.\\
&\left.+\frac{1}{2}\sum_{i,j=1}^{m+n}{(GG^T)_{ij}^{(\theta_p,\theta_v)}\frac{\partial^2 Q^{\theta}}{\partial y_i\partial y_j}}(s_k,a_k)\right)
\end{split}
\end{equation}
as the other obejctive function of APG. Because only $\theta_v$ is optimized in this process, our objective function can be simplified as 
\setcounter{equation}{66}
\begin{equation}
\begin{split}
J'_A(\theta_v)=&-\sum_{k=0}^{l}\left(\sum_{i=1}^{m}{\mu_i^{\theta_v}\frac{\partial Q^{\theta}}{\partial a_i}}(s_k,a_k)\right.\\
&\left.+\frac{1}{2}\sum_{i,j=1}^{m}{(\sigma\sigma^T)_{ij}^{\theta_v}\frac{\partial^2 Q^{\theta}}{\partial a_i\partial a_j}}(s_k,a_k)\right)
\end{split}
\end{equation}
\par
The above two objective functions are derived from the classical policy gradient methods and the framework of IRL. In practice, although the second objective function is more suitable for the overall IRL framework, its optimization process shows some difficulties in converging. The first objective function based on the classical policy gradient also has converging and gradient explosion issues in the IRL training process. How to improve APG's action strategy to make the estimation process more reliable remains an important research question.\par
In the next section, we introduce the algorithm, framework, network structure, flow chart and other technical challenges of IRL.

\section{Technical Frame}\label{sec:Technical Frame}
In the previous sections, we obtained the objective functions (38), (49), (62), and (67), which play a critical role in the network training. There are some technical challenges in achieving IRL: 
\begin{itemize}
	\item How to choose the network model and what problems should be considered when choosing the model?
	\item When the parameter update process is executed, what quantity is required in the parameter update process?
	\item How to reduce the computational complexity?
\end{itemize}
In this section we discuss these technical problems and establishes the framework for IRL.\par

\begin{figure}[!t]
\centering
\includegraphics[width=8.8cm]{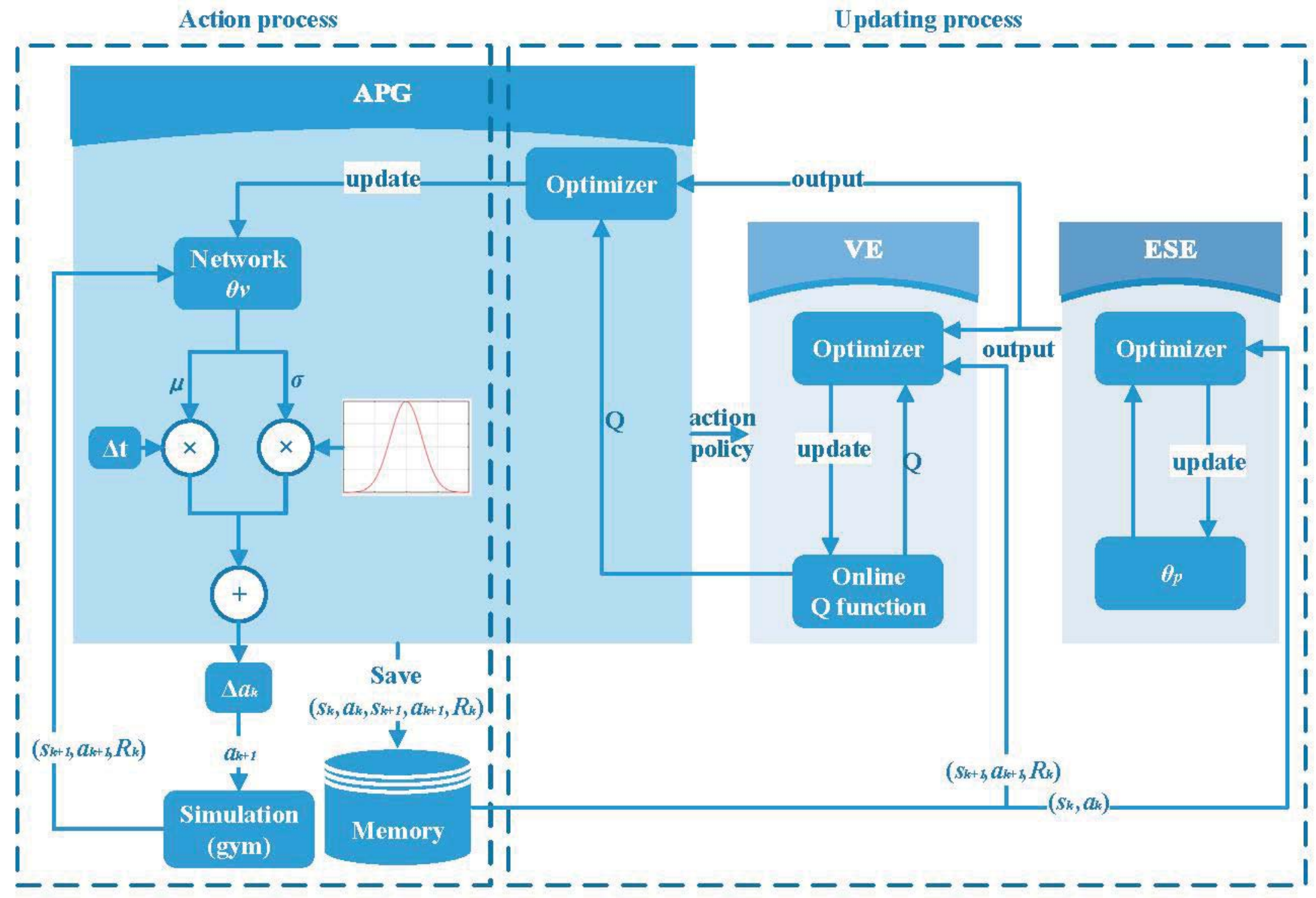}
\caption{The action process and updating process of IRL}
\label{fig: the action process and updating process}
\end{figure}
\floatname{algorithm}{Algorithm}
\renewcommand{\algorithmicrequire}{\textbf{Initialization:}}
\renewcommand{\algorithmicensure}{\textbf{Output:}}

\begin{algorithm}[ht]
	\caption{Incremental Reinforcement learning Algorithm}
	\begin{algorithmic}[1]
		\Require \\
		Randomly initialize network with parameters $\theta$, $\theta_v$, $\theta_p$; target network with parameters $\theta'\xleftarrow{}$ $\theta$,$\theta_v'$ $\xleftarrow{}$ $\theta_v $, $\theta_p'$ $\xleftarrow{}$ $\theta_p'$;replay buffer \it{R};\rm the form of objective function \it{J}$_{Q}$($\theta$),\it{J}$_{E}$($\theta_p$), \it{J}$_{A}$($\theta_v$) \rm as eq(38), (49) ,(62) or (67); max train number $M_{max}$; max train step number $N_{max}$; constant time interval $\Delta t$; dimensions of state and action are m and n respectively; the attenuation rate of future expected reward $\gamma$		
		\While{$M \leq  M_{max}$}
		\State release buffer R
		\State $\partial\theta,\partial\theta_v,\partial\theta_p=0$
		\State $\theta'=\theta,\theta_v'=\theta_v,\theta_p'=\theta_p$
		\State decide the initial state and action $(s,a)$ and $(s_0,a_0)=(s,a)$
		\While{$k\leq N_{max}$}
		\State perform the action $a_k$
		\State recive the reward $R_k$ and $s_{k+1}$
		\State $\Delta B_k\sim N(0,\Delta t \cdot I_n)$
		\State $\Delta a_k=\mu^{\theta_v}(s_k,a_k)\Delta t+\sigma^{\theta_v}(s_k,a_k)\cdot\Delta B_k$
		\State $a_{k+1}=a_k+\Delta a_k$
		\State store $(s_{k+1},a_{k+1})$ and $R_k$ in R
		\State k = k + 1
		\EndWhile	
		\State Calculate $J_{Q}(\theta)$, $J_{E}(\theta_p)$, $J_{A}(\theta_v)$ on the data $\{(s_k,a_k,R_k)\}$ in the Memory
		\State update the parameters$(\theta, \theta_p, \theta_v)$ by $\partial\theta=\partial J_{Q}(\theta), \partial\theta_p=\partial J_{E}(\theta_p),\partial\theta_v=\partial J_{A}(\theta_v)$
		\State sample the train units in the training process and store it in the Memory
		\State M = M + 1
		\EndWhile
		
	\end{algorithmic}
\end{algorithm}
Figure \ref{fig: the action process and updating process} shows the execution procedure of IRL. At each step, the system inputs the current environment state and action state $(s_k, a_k)$, and returns the current increment $\Delta a_k$ inside APG, while ESE and VE are dormant. In this process, it is necessary to store the training data in the memory to update the network by randomly sampling. The training units are $(s_k,a_k,R_k,s_k',a_k')$, where $(s_k',a_k')$ is the next state. The sampling method could be average sampling, or we can first compress the reward into certain interval, (e.g. (0.3, 0.9)), and Bernoulli sampling to decide which training units should be stored in the Memory.\par
\begin{figure*}[ht]
\centering
  \subfigure{\includegraphics[width=5.5cm]{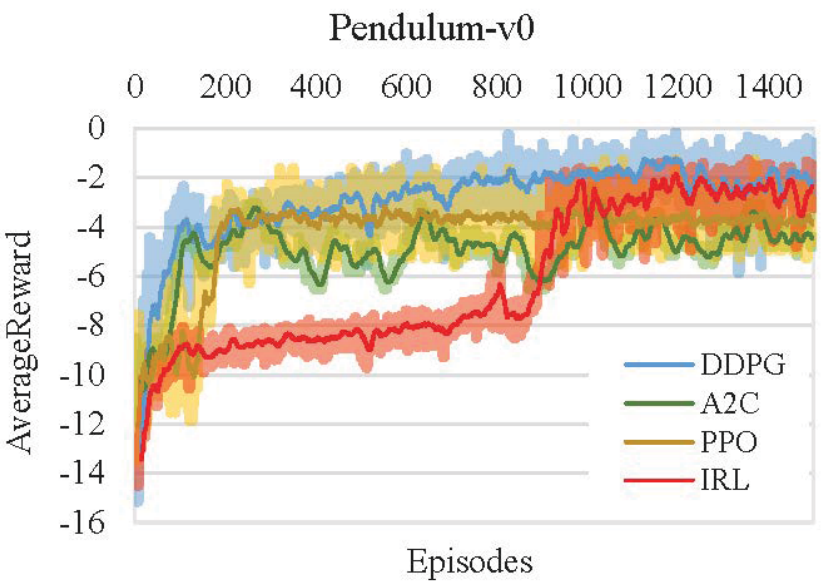}}
  \subfigure{\includegraphics[width=5.5cm]{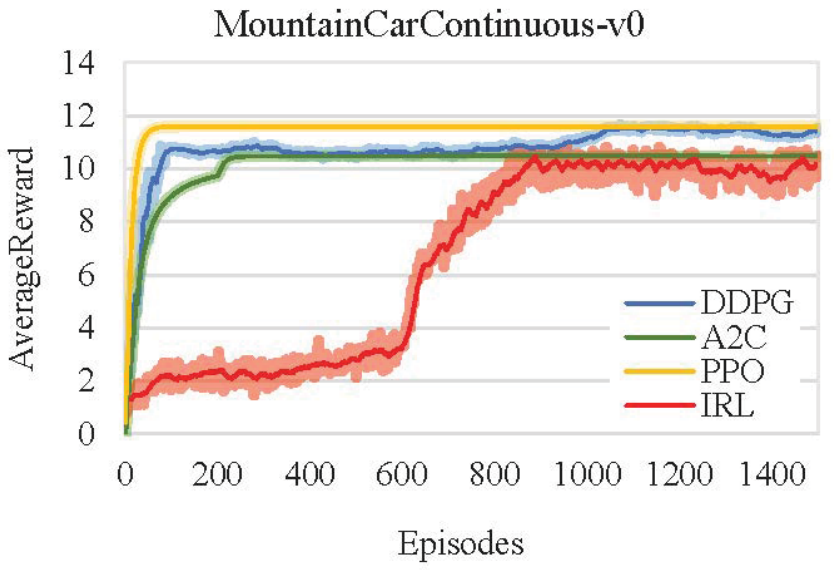}}
  \subfigure{\includegraphics[width=5.5cm]{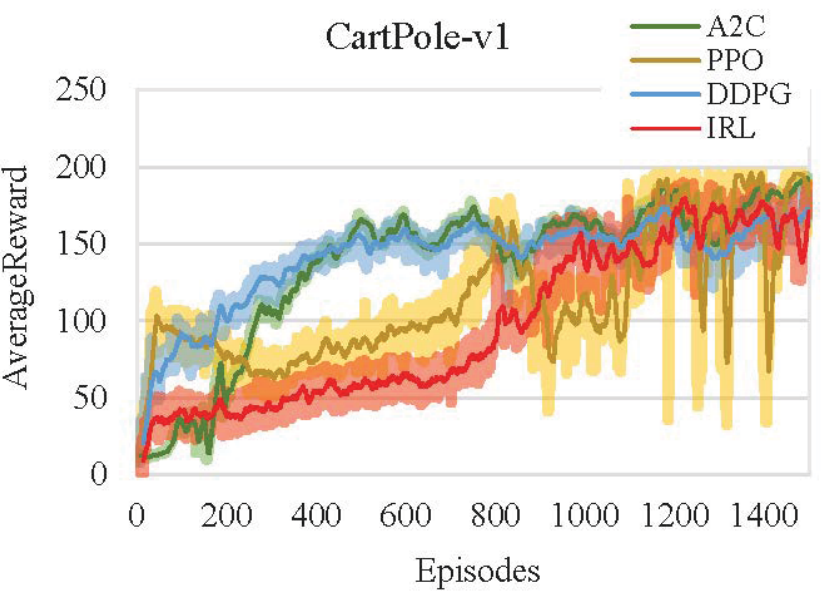}}
\caption{ MountainCarContinuous-v0, Cartpole-v1 and Pendulum-v0 training curves. The training curves in the figure are averaged the three times experiments, and the shaded region shows the standard deviation. }
\label{fig: experiments}
\end{figure*}
Because the network processes time series data, it seems that the network model with hidden variables such as RNN and its variants LSTM, GRU, SRU is suitable for the network structure of IRL. However, they are not ideal choices. As we can see from eq (1) and eq (2), the increment of action and environment follows Markov property, and the output of the network is incremental. If we adopt a recurrent neural network with time series, the incremental generation will no longer follow the Markov property, which is contrary to our model assumptions. The process of estimating the value of Q function in VE is to estimate the expected future revenue. From the deduction process after eq (10), we can see that Q function also satisfies Markov property. Therefore, it is not suitable to use recurrent neural network. In summary, although we deal with time series data in the training process, it is not ideal to use RNN and other recurrent neural networks, so we just use general depth network. On the other hands, we have mentioned earlier that because of the need to ensure the second-order continuity of functions in ESE and VE, the ReLU class of non-linear functions can not be used in the network structure. Otherwise, we can not even calculate the differentials of those objective functions. In the actual computation, we rarely encounter zero points which are the non-differentiable points of ReLU, so we might still use ReLU or PreLU. The possible influence of taking this kind of non-linear function as the non-linear layer of the network is beyond the scope of this paper.\par

The current action output of the agent does not seem to depend on the current state $s_k$, but on the past $(s_ {k-1}, a_ {k-1})$, which means IRL is not Markov control method. Figure \ref{fig: the action process and updating process} shows that the agent does not make actions based on the current environment state, instead they estimates the increment of actions based on the previous environment and action state, which shows their ability to predict under IRL. This ability help adjust its actions continuously. In this situation, agents no longer adapt to the environment to get better rewards. Instead, they adjust according to the environment and its own state. Taking robots in executing tasks for example, when the sensors of the robot transform physical signal and then process the network to make actions, delay occurs, and the external environment of the robot could have changed during the delay. Despite that more advanced sensors, actuators and controllers can largely reduce such a effect, its impacts cannot be ignored. Consequently, the real-time performance of Markov control is always a big challenge. IRL, on the other hand, does not suffer from time delay at all.\par
The pseudo-code for offline IRL can be seen in Algorithm1.
In the next section, we show the performance of IRL, and compare it with other classical continuous reinforcement learning methods.

\section{Experiments Results}\label{sec:Experiment Results}
In this section, we present a comparative experiment that compares IRL and three classical continuous reinforcement learning algorithms (DDPG, A2C and PPO) using three sets of OpenAI Baselines (which are sets of improved implementations of RL algorithms and high-quality implementations of reinforcement learning algorithms \cite{FV:17}). We conduct experiments under three testing environments, Pendulum-v0, MountainCarContinuous-v0 and CartPole-v1 from OpenAI Gym. We record the reward values of each learning method, and plot them against episodes in a episodes-rewards diagram. The goal of our experiments is to verify the feasibility and effectiveness of IRL algorithm.\par
In our experiments, time interval $\Delta t$ is 0.05, $\gamma$ is 0.6, the step number in every epoch is 200. In APG, two kinds of object functions eq(62) and eq(67) jointly update APG network. The weighting coefficient of two objective functions is set to be (0.1,1). In DDPG, A2C and PPO, 200 steps were trained in each episode for a total of 2000 training episodes.\par
As shown in the Figure \ref{fig: experiments}, the experimental effect and performance of IRL basically exceeds those of A2C method, but generally under-performs comparing to DDPG method. Especially in the MountainCarContinuous, IRL is less stable than the other three methods and converges slower. This may be attributed to the fact that the action policy of the network in this experiment is discrete (positively and negatively correlated with the output of the action network) so it is not a strictly continuous control problem and might cause unstability of IRL in the experiment. In the Pendulum-v0 experiment, despite a slower converging speed, IRL reaches same average rewards level as the DDPG. The stability of the dynamic in training process is also quite good. Similar performance is found in CartPole-v1.\par
The reason why our experiments results do not reflect the superiority of the theory might be understood as follows:
\begin{itemize}
\item The updating process of IRL includes partial differential equations as constraints. However, the parameters of the partial differential equations vary during the algorithm updating process. The influence of parameter changes during estimation process due to parameter gradient updates on the constrains requires additional studies on the partial differential equation theory.
\item The convergence of the action network is slower than the value estimation network and the environment estimation network. It indicates that the two action updates used currently are less convergent so the theory on parameter adjustment might need improvement.
\item The current discretization of our control problem may also have some impacts. The theoretical derivation in this paper is based on stochastic differential equation, which is discretized in practical experiments. Other more complex discretization methods may yield different results.
\item One problem with IRL is that it does not constrain the output of an action by adding tanh to the end like DDPG, which might cause boundary overflow problem. In this paper, a boundary penalty regulation term is added
\setcounter{equation}{67}
\begin{equation}
\begin{split}
 J_{Range}(\theta)\!&=\lambda_{Range}\!\sum_{k=0}^{l}\bigg\{\Vert((s_k\!+\!\Delta s_k,a_k\!+\!\Delta a_k )\\
&\!-\!(s_{max},a_{max} ))^+ \Vert_2^2\!+\!\Vert((s_{min},a_{min} )\\
&\!-\!(s_k\!+\!\Delta s_k,a_k\!+\!\Delta a_k ))^+ \Vert_2^2 \bigg\}  
\end{split}
\end{equation}
where + represent ReLU function, $(s_{max},a_{max})$ and $(s_{min},a_{min})$ are the upper and lower bounds of the parameter boundary conditions for the action and the environment. However, it is not clear whether this strategy would affect the convergence of our algorithm.

\end{itemize}\par

\section{Conclusion}\label{sec:Conclusion}
We propose a new continuous reinforcement learning frame IRL. This method guarantees the continuity of actions and the existence and uniqueness of the value estimation network. On one hand, the variance of action strategy is adjusted to control for the randomness of action strategy. On the other hand, an environment estimation network is introduced to improve agents' action adjustments. With our method, agents no longer blindly adapt to the environment but instead estimates the increment of actions based on the previous environment \& action states and make corresponding adjustments continuously. Our method largely reduces the real time delay in the information updates and decision making process so agents' actions are smoother and more accurate.\par

Moreover, the IRL is constructed based on stochastic differential equation methods. Its differential behavior, which is also the important inherent feature of the physical world, makes it suitable for controlling the continuous systems naturally. This behavioral consistency between the target model and the method not only makes the simulation be more realistic than the classical continuous RL methods, such as A3C and DDPG, so as to obtain better continuity and stability, but also obtains better motion correction and control optimization, as shown in figure \ref{fig: The application and the advantages of IRL}. In summary, the proposed IRL provides a new theoretical basis and technical approach for the practical control applications.\par

Our stimulation results are largely consistent with our theory prediction. Although it does not fully reflect the advantage of our theory due to slow convergence issues, we have identified some of the limitations in our experiment, so in the future research we will experiment on using partial differential equation theory to estimate Q function and improve the convergence speed during optimization. We will also solve the boundary overflow problem.\par 
\begin{figure}[t]
\centering
\includegraphics[width=8cm]{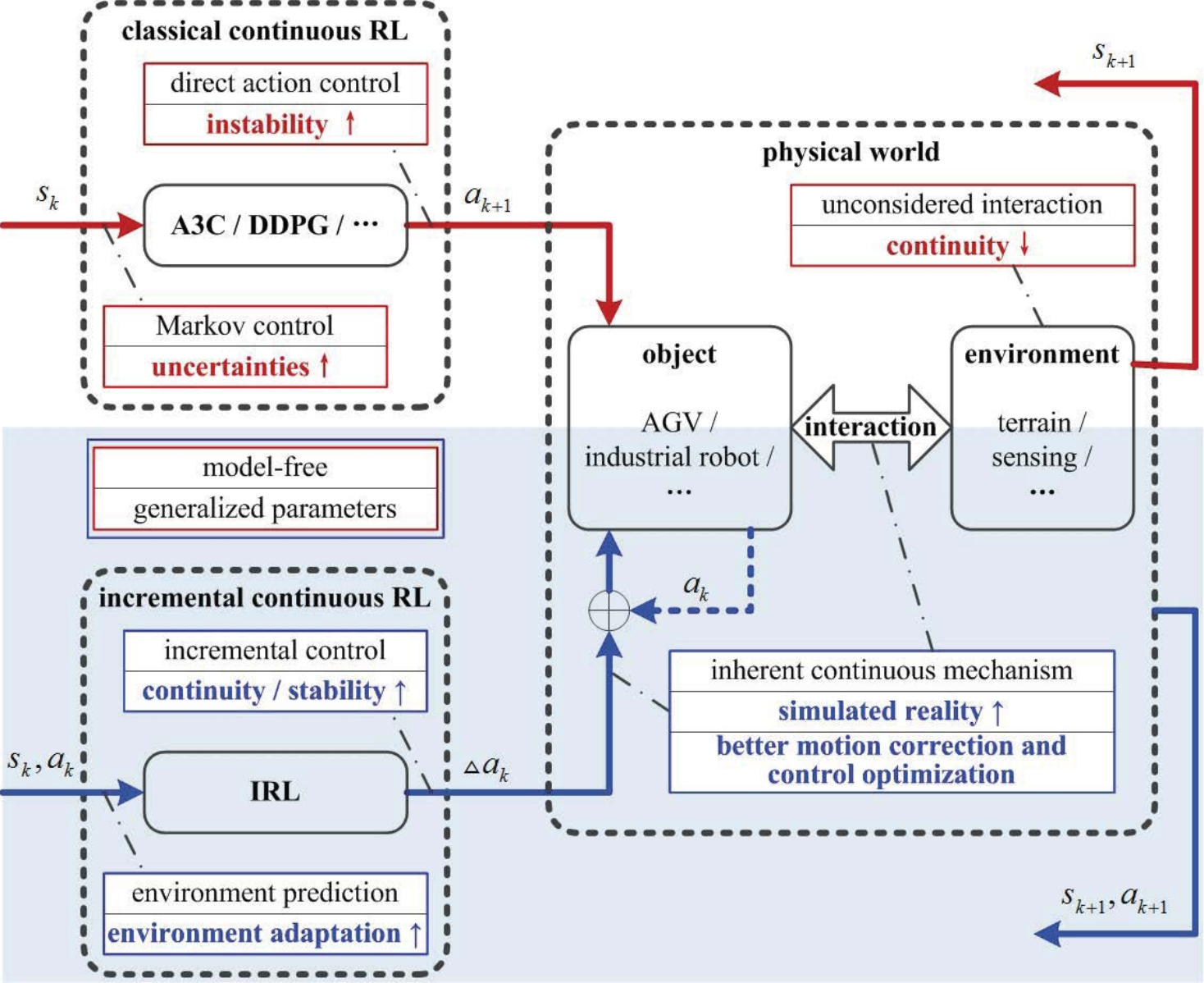}
\caption{The application and the advantages of IRL}
\label{fig: The application and the advantages of IRL}
\end{figure}

\newpage
\appendix
\appendices
\label{app:theorem}
In the appendix we prove that the action policy of some classical reinforcement learning methods such as A3C cannot guarantee the continuity of action, which means those methods cannon be applied in many tasks such as robot self-adaptive control. We start with a definition of the modification of stochastic process.
\noindent

{\bf Definition} {\it Suppose process $\{X_t\}$ and $\{Y_t\}$ are both the stochastic processes on $(\Omega,\mathcal{F},P)$. If
	$$P(\{\omega;X_t(\omega)=Y_t(\omega)\})=1\ \forall t$$
then $\{Y_t\}$ is a modification of $\{X_t\}$, and they will have the same finite dimensional distribution.
} \hfill
\noindent

Now give an important continuity theorem -- the Kolmogorov continuity theorem.
\noindent

{\bf Theorem} {\it Suppose process $X=\{X_t\}_{t\geq 0}$ meets the conditions: $T>0$, if there exists positive constants $m,n,D$ such that
	$$E\left[|X_t-X_s|^m\right]\leq D\cdot |t-s|^{1+n};\ 0\leq s,t\leq T$$
then there exists a continuous modification of $X$.
} \hfill

Now let's prove that AC method using Gauss policy as action policy can't satisfy the conditions for continuity of agents' actions. The general form of this action policy is
$$a_t = \mu(s_t)+\sigma(s_t)\cdot \mathcal{N}_t$$
where $\mathcal{N}_t$ is white noise. Suppose ${s_t}_{\geq 0}$ are independent of each other at any time and $a_t$ is one-dimensional. 
\noindent

{\bf Proof}. $a_t$ follows Gauss distribution, that $a_t\!\sim\! N(\mu(s_t),\sigma^2(s_t))$, and 
$$a_t-a_p\sim N(\mu(s_t)-\mu(s_p),\sigma^2(s_t)+\sigma^2(s_p))$$
for all $m>0$, the moment conditions of absolute values can be expressed in terms of confluent hypergeometric functions:
\begin{equation}
\begin{split}
E\left[|a_t-a_p|^m\right]\!=\!&\left(\sigma^2(s_t)+\sigma^2(s_p)\right)^{\frac{m}{2}}\cdot 2^{\frac{m}{2}}\frac{\Gamma(\frac{1+m}{2})}{\sqrt{\pi}}\ _1\\
&\cdot F_1 \left(-\frac{m}{2},\frac{1}{2},-\frac{1}{2}\frac{(\mu(s_t)-\mu(s_p)^2)}{\sigma^2(s_t)+\sigma^2(s_p)}\right)
\end{split}
\end{equation}
where $\Gamma$ is the Gamma function and $\ _1F_1$ is the Confluent hypergeometric function:
$$\ _1F_1(a,b,z)=\sum_{n=0}^{\infty}\frac{a^{(n)}z^n}{b^{(n)}n!}$$
where$a^{(0)}=1,a^{(n)}=a(a+1)...(a+n-1)$. It is easy to see there exists a positive $\left(\sigma^2(s_t)+\sigma^2(s_p)\right)^{\frac{m}{2}}\cdot 2^{\frac{m}{2}}
\frac{\Gamma(\frac{1+m}{2})}{\sqrt{\pi}}$ that is irrelevant to $|t-s|$ even with $\mu$ satisfying 
H\"{o}lder condition, so $a_t$ cannot satisfy Kolmogorov continuity and it doesn't exists a continuous modification.

\footnotesize
\itemsep=-3pt plus.2pt minus.2pt
\bibliographystyle{IEEEtran}
\bibliography{IEEEabrv,IRL}
\vfill
\enlargethispage{-5in}

\else
\fi
\end{document}